\documentclass{article}

\usepackage{arxiv}
\usepackage[utf8]{inputenc}
\usepackage[T1]{fontenc}
\usepackage{hyperref}
\usepackage{url}
\usepackage{booktabs}
\usepackage{amsmath}
\usepackage{amssymb}
\usepackage{amsfonts}
\usepackage{nicefrac}
\usepackage{microtype}
\usepackage{graphicx}
\usepackage[table]{xcolor}
\usepackage{tikz}
\usetikzlibrary{arrows.meta,calc,fit,positioning}
\usepackage{xspace}
\usepackage{flafter}
\usepackage{placeins}

\graphicspath{{./figures/}}

\title{Sobek: Streaming Equivariant Tensor Product Convolutions}

\author{
  Vladimir Choro\v{s}ajev \\
  Cortex Discovery \\
  \texttt{vladimir.chorosajev@cortexdiscovery.com} \\
  \And
  Cédric Bény \\
  Cortex Discovery \\
  \texttt{cedric.beny@cortexdiscovery.com}
}

\newcommand{\R}{\mathbb{R}}
\newcommand{\E}{\mathcal{E}}
\newcommand{\V}{\mathcal{V}}
\newcommand{\N}{\mathcal{N}}
\newcommand{\sobek}{\textsc{sobek}\xspace}
\definecolor{sobekblue}{HTML}{0072B2}
\definecolor{openeqorange}{HTML}{D55E00}

\newcommand{\CG}{\mathrm{CG}}
\newcommand{\scinum}[2]{\ensuremath{#1\times10^{#2}}}

\begin{document}
\raggedbottom
\maketitle

\begin{abstract}
Equivariant graph neural networks repeatedly apply edge-conditioned
tensor-product convolutions over graph edges.  Conventional implementations
materialize edge-specific weights, messages, and adjoints, causing
tensor-product workspace and memory traffic to grow rapidly with graph size
and operator width.  This limits feasible workloads and can prevent larger
problems from fully utilizing the GPU.

We show that these edge-sized intermediates are artifacts of the execution
schedule, not requirements of the equivariant operator.  By reassociating
radial projection, spherical-harmonic coupling, and graph aggregation,
edge-local products can be consumed directly into bounded receiver-side state.
The resulting streaming formulation preserves fully connected multiplicity
mixing and extends through forward, backward, and double backward.

We implement this formulation in \sobek \footnote{Source: \url{    https://github.com/Cortex-Discovery/sobek}}, a generated-CUDA backend, and evaluate
it across edge-scaling regimes and varied feature structures.  Across two
operator families and all three differentiation orders, \sobek is faster in all
75 capacity-matched comparisons, with speedups ranging from $1.2\times$ to
$49.7\times$, and reduces peak allocated memory by up to 99\%.  It also
executes workloads up to two orders of magnitude beyond OpenEquivariance's
capacity while retaining near-peak throughput.  These results show that
edge-scaled tensor-product workspace is a property of the conventional
schedule, not of equivariant convolution itself.
\end{abstract}


\section{Introduction}

Geometric deep learning encodes symmetries in model layers, extending translation- and permutation-aware architectures to rotations, reflections,
and translations in three dimensions~\cite{bronstein2021geometric}.
Equivariant models have been successfully applied to spherical images (Clebsch--Gordan Nets~\cite{kondor2018cgnets}), point clouds and N-body simulation (SE(3)-Transformers~\cite{fuchs2020se3transformer}), and computational
physics (SEGNNs~\cite{brandstetter2022segnn}).

Among equivariant architectures, one influential family represents node
features as irreducible representations ({\em irreps}), distances through radial
basis embeddings, and directions through spherical harmonics.  Clebsch--Gordan
tensor products combine node features with radial and angular edge features
while preserving their prescribed transformations.  Tensor Field Networks
established this construction for point-cloud
convolutions~\cite{thomas2018tfn}; we refer to the resulting operator primitive
as \emph{edge-conditioned geometric convolution}.

This operator primitive has become especially prominent in chemical machine
learning, where machine-learned interatomic potentials must
predict invariant energies and equivariant forces from expensive reference
data.  NequIP demonstrated the data efficiency and accuracy of symmetries as inductive
bias~\cite{batzner2022nequip}; MACE, Equiformer, and Allegro likewise use this
geometric primitive for molecular and materials
modeling~\cite{batatia2022mace,liao2023equiformer,musaelian2023allegro}.

The operator is computationally and memory intensive.  Each edge couples sender
features, spherical harmonics, and edge-specific weights across tensor-product
paths.  Cost grows with edge count: wider representations enlarge channel
contractions, higher angular order adds coupling paths and component arithmetic,
and larger batches and cutoffs multiply edges.  Tensor-product convolution can
therefore dominate training and inference, forcing tradeoffs among width,
angular order, interaction range, and batch size.

This cost can constrain not only model scale, but also the expressivity of the
geometric convolution itself.  Practical architectures may restrict how input
and output channels interact to keep edge-conditioned weights tractable
\cite{liao2023equiformer}.  These channelwise or depthwise operators can be
effective, but they represent a narrower family of edge-conditioned
equivariant maps than fully connected mixing~\cite{xie2025price}.

Systems such as OpenEquivariance~\cite{bharadwaj2025openequivariance} and
NVIDIA cuEquivariance~\cite{nvidia2025cuequivariance} accelerate tensor-product
evaluation through specialized kernels.  Architectural approaches such as
EquiformerV2~\cite{liao2024equiformerv2} instead reduce the arithmetic cost of
the angular operator.  The present work addresses a different question: the
dataflow of the complete edge-conditioned convolution.

A conventional decomposition first expands each graph edge embedding into a vector
of tensor-product weights, applies the tensor product to form an edge message,
and finally aggregates messages at receiver nodes.  The expanded weights and
messages are single-use intermediates, yet their storage scales with the number
of edges.  They can therefore dominate peak memory and memory traffic for wide
representations or high-density geometric graphs.

We show that these edge-sized tensors are scheduling artifacts, derive a
streaming schedule by reassociating the contractions as receiver-owned
reductions, and introduce \sobek as a generated-CUDA backend that implements
this schedule for edge-conditioned equivariant convolutions.
\sobek never materializes complete edge-weight or edge-message tensors while
preserving the operator's forward, backward, and double-backward semantics.
The reformulation is algebraically equivalent to the conventional schedule,
and every reported measurement passes strict double-peer numerical validation.
Only inherent graph and feature storage remains edge-scaled; tensor-product
workspace is bounded by a fixed scratch budget.  The resulting implementation
also delivers higher sustained throughput and permits larger workloads to
occupy the GPU more fully.

This work makes three contributions:

\begin{itemize}

\item \textbf{Algorithm:} We identify per-edge weights, messages, and adjoints
as scheduling artifacts and derive algebraically equivalent streamed forward,
backward, and double backward with bounded tensor-product workspace.

\item \textbf{System:} We provide an implementation of the streaming schedule
in \sobek, a deterministic CUDA code-generation backend supporting fully
connected \texttt{uvw} and channelwise \texttt{uvu} execution across all three
differentiation orders without edge-weight workspace.

\item \textbf{Evidence:} Across controlled scaling sweeps and 25 feature
profiles, \sobek is faster in all 75 capacity-matched comparisons, reduces peak
allocation by up to 99\%, and sustains near-peak throughput at workloads far
beyond OpenEquivariance's capacity.

\end{itemize}

\section{Background}

\subsection{Geometric equivariance}

Geometric equivariance specifies how model outputs transform when the input
geometry is transformed.  Invariant outputs remain unchanged, while
equivariant outputs follow a prescribed transformation law.  Molecular energy
prediction provides a concrete example: for a rotation matrix $R$ and atomic
positions $\mathbf r$, energy remains unchanged while forces rotate with the
positions:
\begin{equation}
  E(R\mathbf r)
  =
  E(\mathbf r),
  \qquad
  \mathbf F(R\mathbf r)
  =
  R\mathbf F(\mathbf r).
  \label{eq:energy_force_symmetry}
\end{equation}

More generally, let $\pi_{g}$ and $\pi_{g}'$ describe how a transformation $g$ acts respectively on the input and output of a map $f$. We say that $f$ is equivariant when
\begin{equation}
  f\!\left(\pi_{g}(x)\right)
  =
  \pi_{g}'(f(x)).
  \label{eq:equivariance_definition}
\end{equation}


Linear actions of the group of improper rotations $\mathrm{O}(3)$ can be expressed as a direct sum of irreducible representations (irreps). Here we consider the case where $f$ is a bilinear map, and $\pi_g$ is a tensor product of linear representations. Any such equivariant $f$ is a linear combination of basic operations (paths) characterized by the Clebsch-Gordon coefficients, and mapping specific input irreps into a specific output irrep. In geometric GNNs, each edge of the graph is associated to such an operation, with edge-dependent path weights~\cite{thomas2018tfn,geiger2022e3nn}.

\subsection{Irreducible features and tensor products}

An equivariant feature field is a direct sum of irreducible representations,
\begin{equation}
  x
  =
  \bigoplus_{\ell,\sigma}
  x_{\ell,\sigma},
  \qquad
  x_{\ell,\sigma}
  \in
  \R^{C_{\ell,\sigma} \times (2\ell+1)},
\end{equation}
where $\ell$ is the angular order, $\sigma\in\{-1,1\}$ is parity, and
$C_{\ell,\sigma}$ is the multiplicity of that irrep.  The group acts on the
$2\ell+1$ angular components, while multiplicities form a learned channel
axis.

The flattened feature dimension is
\begin{equation}
  C_x
  =
  \sum_{\ell,\sigma} C_{\ell,\sigma}(2\ell+1).
  \label{eq:flattened_feature_dimension}
\end{equation}

Tensor-product layers combine an input irrep with a spherical-harmonic irrep.
A valid output satisfies the Clebsch--Gordan selection rules
\begin{equation}
  \ell_{\mathrm{out}}
  \in
  |\ell_{\mathrm{in}}-\ell_{\mathrm{sh}}|,
  \ldots,
  \ell_{\mathrm{in}}+\ell_{\mathrm{sh}},
  \qquad
  \sigma_{\mathrm{out}}
  =
  \sigma_{\mathrm{in}}\sigma_{\mathrm{sh}}.
  \label{eq:cg_selection_rules}
\end{equation}
Each valid combination defines a tensor-product path.  Clebsch--Gordan
coefficients determine how its angular components couple, while learned
weights act only on multiplicities. 

\subsection{Channelwise and fully connected multiplicity mixing}

Tensor-product connection modes specify how features are mixed along the multiplicity axis for a path allowed by Clebsch-Gordan selection rules~\cite{geiger2022e3nn,e3nn2026tensorproductdocs}.  Let $u$, $v$, and
$w$ index the input, spherical-harmonic, and output multiplicities of one valid
angular path.  For an edge $e$ with unit direction $\widehat{\mathbf r}$, a fully connected \texttt{uvw}
instruction computes
\begin{equation}
  z_w
  =
  \sum_{u,v}
  \omega_{uvw}(e)
  \left[x_u \otimes_{\CG} Y_v(\widehat{\mathbf r})\right],
  \label{eq:uvw_connection}
\end{equation}
and therefore uses
$C_{\mathrm{in}} \cdot C_{\mathrm{sh}} \cdot C_{\mathrm{out}}$ edge-dependent weights for
that path.  In the channelwise \texttt{uvu} mode, the repeated index ties the
output multiplicity to the first input multiplicity,
\begin{equation}
  z_u
  =
  \sum_v
  \omega_{uv}(e)
  \left[x_u \otimes_{\CG} Y_v(\widehat{\mathbf r})\right],
  \label{eq:uvu_connection}
\end{equation}
so the instruction requires $C_{\mathrm{out}}=C_{\mathrm{in}}$ and only
$C_{\mathrm{in}}C_{\mathrm{sh}}$ weights.  This equality is local to one
instruction; different irrep blocks may still have different multiplicities,
and later equivariant linear layers may mix their outputs.

These modes differ in their expressivity.  A fully connected \texttt{uvw} path allows the invariant edge
condition---in molecular models, typically a function of interatomic
distance---to select a different mixing between every input and output
multiplicity channel.  A \texttt{uvu} path can still modulate each tied channel
with distance, but it cannot change which input and output channels are coupled
within that path.  Surrounding equivariant linear layers can still produce
expressive networks, but they do not make one \texttt{uvu} instruction
equivalent to a general \texttt{uvw} instruction at fixed width.  This operator-level expressivity distinction was studied by Xie et
al.~\cite{xie2025price}.

\subsection{Edge-conditioned tensor product convolution}

Let $\mathcal G=(\V,\E)$ be a directed graph with source node $j$ and receiver
node $i$ on edge $j\rightarrow i$.  For node positions $\mathbf r_i$, define
the relative displacement, distance, and direction as
\begin{equation}
  \mathbf r_{ij}=\mathbf r_j-\mathbf r_i,
  \qquad
  r_{ij}=\lVert\mathbf r_{ij}\rVert,
  \qquad
  \widehat{\mathbf r}_{ij}=\frac{\mathbf r_{ij}}{r_{ij}}.
\end{equation}
The angular input $Y_{ij}$ contains spherical harmonics of
$\widehat{\mathbf r}_{ij}$, while $\phi_{ij}\in\R^K$ is a radial or more
general invariant edge embedding.

Let $\mathcal P$ be the set of valid tensor-product paths.  Each path
$\pi\in\mathcal P$ selects an input irrep, a spherical-harmonic irrep, an output
irrep, and their Clebsch--Gordan coupling.  Because the spherical harmonics have
multiplicity one, path $\pi$ has
\begin{equation}
  n_\pi
  =
  C^{\mathrm{in}}_\pi C^{\mathrm{out}}_\pi,
\end{equation}
where $C^{\mathrm{in}}_\pi$ and $C^{\mathrm{out}}_\pi$ are the coupled input
and output multiplicities.  The index $a\in\{1,\ldots,n_\pi\}$ below is a
flattened input/output multiplicity pair.

The total number of edge-specific tensor-product weights is
\begin{equation}
  N_{\mathrm{w}}
  =
  \sum_{\pi \in \mathcal{P}} n_\pi.
  \label{eq:num_weights}
\end{equation}

An edge-conditioned equivariant convolution computes
\begin{equation}
  m_i
  =
  \sum_{j \in \N(i)}
  \sum_{\pi \in \mathcal{P}}
  \sum_{a=1}^{n_\pi}
  W_{ij,\pi,a}
  \;
  T_{\pi,a}
  \left(
    x_j,
    Y_{ij}
  \right),
  \label{eq:standard_tpconv_background}
\end{equation}
where $T_{\pi,a}(x_j,Y_{ij})$ is the angular tensor-product contribution for
path $\pi$ and multiplicity pair
$a$~\cite{batzner2022nequip,batatia2022mace,geiger2022e3nn}, and
$C_{\mathrm{out}}$ denotes the flattened output feature dimension.

The edge-dependent weights are generated by a learned radial projection:
\begin{equation}
  W_{ij,\pi,a}
  =
  \sum_{q=1}^{K}
  \phi_{ij,q} A_{q,\pi,a}.
  \label{eq:radial_projection_background}
\end{equation}
After flattening $(\pi,a)$, the three tensors have shapes
\begin{equation}
  \phi
  \in
  \R^{|\E|\times K},
  \qquad
  A
  \in
  \R^{K\times N_{\mathrm w}},
  \qquad
  W
  \in
  \R^{|\E|\times N_{\mathrm w}},
  \qquad
  W=\phi A.
  \label{eq:edge_weight_shape}
\end{equation}
The model stores $A$ once per layer, and the graph supplies $\phi$ once per
edge.  The tensor $W$ is neither a model parameter nor an input; it is a
transient expansion produced by the radial projection and consumed by the
tensor-product contraction.

\subsection{Memory cost of materialized execution}

A conventional materialized evaluation forms both the edge weights and the
edge messages before reducing them to the node output
$m\in\R^{|\V|\times C_{\mathrm{out}}}$.  For a scalar type with $b$ bytes,
the edge-weight tensor $W$ and edge-message tensor
$M\in\R^{|\E|\times C_{\mathrm{out}}}$ require $b|\E|N_{\mathrm w}$ and
$b|\E|C_{\mathrm{out}}$ bytes, respectively.

The graph inputs $\phi$ and $Y$ are also edge-sized, and any exact execution
must either store or recompute them.  Their widths are $K$ and
$\dim(\mathcal R_{\mathrm{sh}})$, while materialization expands each edge to
$N_{\mathrm w}$ weights and a $C_{\mathrm{out}}$-dimensional message.  The
avoidable cost is therefore not edge dependence itself, but the much wider
intermediates created from those inputs.

Both factors in $|\E|N_{\mathrm w}$ can grow.  At fixed angular structure,
scaling every input and output multiplicity by $s$ gives
\begin{equation}
  N_{\mathrm w}^{\texttt{uvw}}(sC)
  =
  s^2 N_{\mathrm w}^{\texttt{uvw}}(C),
  \label{eq:uvw_width_scaling}
\end{equation}
because each path connects every input multiplicity to every output
multiplicity.  The edge count satisfies $|\E|=|\V|\bar d$, where $\bar d$ is the
average directed degree.  Increasing the number of nodes or their average
degree therefore increases the materialized cost, which reaches
$O(|\V|^2N_{\mathrm w})$ in the dense limit.

For three-dimensional radius graphs at fixed particle density $\rho$,
$|\E|\propto|\V|\rho r_{\mathrm c}^{3}$, where $r_{\mathrm c}$ is the
interaction cutoff.  Work and materialized edge memory therefore grow cubically
with the cutoff.  Table~\ref{tab:example_weight_counts} illustrates how
increasing multiplicities expand materialized edge-weight memory at fixed graph
size.

\begin{table}[!htbp]
  \centering
  \caption{Edge-specific tensor-product weights for equal-width profiles at
  one million directed edges.  Counts assume $L_{\max}=2$, matching
  input/output irreps, one spherical-harmonic copy per angular order, and fully
  connected multiplicity mixing.  Profile entries list multiplicities in
  ascending angular order.}
  \label{tab:example_weight_counts}
  \scriptsize
  \setlength{\tabcolsep}{2.7pt}
  \renewcommand{\arraystretch}{1.08}
  \begin{tabular}{@{}lrr@{}}
    \toprule
    Profile & $N_{\mathrm w}$ & fp32 $M_W$ at \scinum{1.0}{6} edges (GiB) \\
    \midrule
    $32/32/32$    & 11,264  & 42.0 \\
    $64/64/64$    & 45,056  & 167.8 \\
    $128/128/128$ & 180,224 & 671.4 \\
    $256/256/256$ & 720,896 & 2,685.5 \\
    \bottomrule
  \end{tabular}
\end{table}
\FloatBarrier

This memory cost is not inherent to equivariant convolution.  The next section
shows that the same operator, including its first and second derivatives, can
be evaluated without ever assembling the complete edge-weight or edge-message
tensors.

\section{Streaming Execution Schedule}
\label{sec:streaming_schedule}

Streaming is a property of intermediate lifetime, not a particular kernel
architecture.  The materialized evaluation order creates complete tensors $W$
and $M$ in memory, that the convolution does not require.  A streaming schedule instead
consumes their values incrementally without retaining the complete edge-sized
tensors as global intermediates.  This does not require a single fused kernel,
zero scratch, a specific graph ordering, or a particular reduction strategy.

\subsection{Eliminating materialized edge weights}

Substituting the radial projection in
Eq.~\ref{eq:radial_projection_background} into
Eq.~\ref{eq:standard_tpconv_background} gives
\begin{equation}
  m_i
  =
  \sum_{j \in \N(i)}
  \sum_{\pi \in \mathcal{P}}
  \sum_{a=1}^{n_\pi}
  \left(
    \sum_{q=1}^{K}
    \phi_{ij,q} A_{q,\pi,a}
  \right)
  T_{\pi,a}(x_j, Y_{ij}).
  \label{eq:materialized_expression}
\end{equation}

The equivalent streaming order is
\begin{equation}
  m_i
  =
  \sum_{j \in \N(i)}
  \sum_{\pi \in \mathcal{P}}
  \sum_{a=1}^{n_\pi}
  \sum_{q=1}^{K}
  \phi_{ij,q}
  A_{q,\pi,a}
  T_{\pi,a}(x_j, Y_{ij}).
  \label{eq:streaming_expression}
\end{equation}

The two expressions are identical in exact arithmetic.  Both use the same
paths, radial model, and Clebsch--Gordan coefficients to define the same
equivariant operator.  Only the lifetime of intermediate values changes.  Their
accumulation orders may differ in floating-point arithmetic, so numerical
comparisons should use an explicit tolerance.

Define the edge-specific scalar weight as

\begin{equation}
  w_{ij,\pi,a}
  =
  \sum_{q=1}^{K}\phi_{ij,q}A_{q,\pi,a}
  \label{eq:streamed_edge_weight}
\end{equation}

A streaming schedule need only keep this scalar live until its tensor-product
contribution has been consumed.  The full collection
$W\in\R^{|\E|\times N_{\mathrm w}}$ is therefore an artifact of one evaluation
order, not a mathematical input or output of the convolution.  Applying the
same lifetime rule to the contribution
$w_{ij,\pi,a}T_{\pi,a}(x_j,Y_{ij})$ also avoids a complete edge-message tensor.

The lifetime argument is independent of the angular tensor-product algorithm
and multiplicity connection mode: those choices determine arithmetic and
channel connectivity, while streaming determines whether their
edge-conditioned values are retained across the complete graph.

\subsection{Backward without an edge-weight-adjoint tensor}

Let $\bar m_i$ denote the output adjoint and define the edge-local scalar

\begin{equation}
  h_{ij,\pi,a}
  =
  \left\langle
    \bar m_i,
    T_{\pi,a}(x_j,Y_{ij})
  \right\rangle.
  \label{eq:streamed_weight_adjoint}
\end{equation}

The radial-parameter and edge-embedding adjoints are then

\begin{align}
  \bar A_{q,\pi,a}
  &=
  \sum_i\sum_{j\in\N(i)}
  \phi_{ij,q}h_{ij,\pi,a}, \\
  \bar\phi_{ij,q}
  &=
  \sum_{\pi,a}
  A_{q,\pi,a}h_{ij,\pi,a}.
  \label{eq:streamed_radial_adjoints}
\end{align}

For the tensor-product operands, let $D_xT_{\pi,a}^{*}$ and
$D_YT_{\pi,a}^{*}$ denote the adjoints of the two partial Jacobians.  Their
gradients are

\begin{align}
  \bar x_j
  &=
  \sum_{i:\,j\in\N(i)}\sum_{\pi,a}
  w_{ij,\pi,a}
  D_xT_{\pi,a}(x_j,Y_{ij})^{*}\bar m_i, \\
  \bar Y_{ij}
  &=
  \sum_{\pi,a}
  w_{ij,\pi,a}
  D_YT_{\pi,a}(x_j,Y_{ij})^{*}\bar m_i.
  \label{eq:streamed_operand_adjoints}
\end{align}

Every quantity on the right-hand sides can be formed for one edge and path,
accumulated into its destination, and discarded.  The scalar $h_{ij,\pi,a}$
equals the corresponding entry of the edge-weight adjoint but can be consumed
immediately, so the complete tensor
$\bar W\in\R^{|\E|\times N_{\mathrm w}}$ is never needed.  First backward
therefore admits the same streaming schedule as forward.

\subsection{Double backward by differentiating the streamed contractions}

The same locality extends to second order.  Square brackets below denote
application of a partial Jacobian to a direction,
not indexing: $D_xT_{\pi,a}[\dot x_j]$ is the directional derivative with
respect to $x_j$ along $\dot x_j$, and analogously for
$D_YT_{\pi,a}[\dot Y_{ij}]$.

For directions $\dot x$, $\dot Y$, $\dot\phi$, and $\dot A$, define

\begin{align}
  \dot w_{ij,\pi,a}
  &= \sum_q
  \left(
    \dot\phi_{ij,q}A_{q,\pi,a}
    + \phi_{ij,q}\dot A_{q,\pi,a}
  \right), \\
  \dot T_{ij,\pi,a}
  &= D_xT_{\pi,a}[\dot x_j]
  + D_YT_{\pi,a}[\dot Y_{ij}].
  \label{eq:streamed_directional_terms}
\end{align}

The forward directional output and the directional edge-weight adjoint are

\begin{align}
  \dot m_i
  &= \sum_{j\in\N(i)}\sum_{\pi,a}
  \left(
    \dot w_{ij,\pi,a}T_{\pi,a}(x_j,Y_{ij})
    + w_{ij,\pi,a}\dot T_{ij,\pi,a}
  \right), \\
  \dot h_{ij,\pi,a}
  &= \left\langle\dot{\bar m}_i,
    T_{\pi,a}(x_j,Y_{ij})\right\rangle
  + \left\langle\bar m_i,\dot T_{ij,\pi,a}\right\rangle.
  \label{eq:streamed_second_order_local_terms}
\end{align}

Both depend only on edge-local values.  Differentiating the first-order
adjoints applies the same product rule; for example,

\begin{align}
  \dot{\bar A}_{q,\pi,a}
  &= \sum_i\sum_{j\in\N(i)}
  \left(
    \dot\phi_{ij,q}h_{ij,\pi,a}
    + \phi_{ij,q}\dot h_{ij,\pi,a}
  \right), \\
  \dot{\bar\phi}_{ij,q}
  &= \sum_{\pi,a}
  \left(
    \dot A_{q,\pi,a}h_{ij,\pi,a}
    + A_{q,\pi,a}\dot h_{ij,\pi,a}
  \right).
  \label{eq:streamed_radial_second_adjoints}
\end{align}

For the operand adjoints, define the local contractions

\begin{equation}
  B^x_{ij,\pi,a}=D_xT_{\pi,a}^{*}\bar m_i,
  \qquad
  B^Y_{ij,\pi,a}=D_YT_{\pi,a}^{*}\bar m_i,
\end{equation}

and let $\dot B^x$ and $\dot B^Y$ denote their directional derivatives.  The
remaining second adjoints are

\begin{align}
  \dot{\bar x}_j
  &= \sum_{i:\,j\in\N(i)}\sum_{\pi,a}
  \left(
    \dot w_{ij,\pi,a}B^x_{ij,\pi,a}
    + w_{ij,\pi,a}\dot B^x_{ij,\pi,a}
  \right), \\
  \dot{\bar Y}_{ij}
  &= \sum_{\pi,a}
  \left(
    \dot w_{ij,\pi,a}B^Y_{ij,\pi,a}
    + w_{ij,\pi,a}\dot B^Y_{ij,\pi,a}
  \right).
  \label{eq:streamed_operand_second_adjoints}
\end{align}

Because the tensor product is bilinear, $\dot B^x$ and $\dot B^Y$ use the same
edge operands and Clebsch--Gordan coefficients as the forward and first-order
contractions.  The complete second-order program therefore remains edge-local
and requires no materialized $W$, $\bar W$, directional edge-weight tensor,
edge-message tensor, or dense Hessian.

\subsection{Global lifetime summary}

Table~\ref{tab:memory_complexity} summarizes the complete global tensors used
by materialized and streaming schedules.  It does not constrain local or
bounded workspace within a particular implementation.

\begin{table}[!htbp]
  \centering
  \caption{Complete global tensors used by materialized and streaming
  tensor-product convolution schedules.}
  \label{tab:memory_complexity}
  \scriptsize
  \setlength{\tabcolsep}{2.7pt}
  \renewcommand{\arraystretch}{1.08}
  \begin{tabular}{@{}llcc@{}}
    \toprule
    Quantity & Shape & Materialized & Streaming \\
    \midrule
    Node features
      & $|\V| \times C_{\mathrm{in}}$
      & stored
      & stored \\
    Edge embeddings
      & $|\E| \times K$
      & stored
      & stored \\
    Spherical harmonics
      & $|\E| \times \dim(\mathcal R_{\mathrm{sh}})$
      & stored or recomputed
      & stored or recomputed \\
    Edge TP weights
      & $|\E| \times N_{\mathrm{w}}$
      & materialized
      & not required as a complete tensor \\
    Edge messages
      & $|\E| \times C_{\mathrm{out}}$
      & materialized or partially materialized
      & not required as a complete tensor \\
    Node outputs
      & $|\V| \times C_{\mathrm{out}}$
      & stored
      & stored \\
    \bottomrule
  \end{tabular}
\end{table}
\FloatBarrier

The transformation is therefore conceptually independent of data layout,
precision, tensor-product instruction representation, and hardware mapping.
The next section shows how this implementation-independent principle can be
realized as a practical execution architecture without weakening the
edge-tensor lifetime guarantee.

\section{\texorpdfstring{\sobek}{sobek} Design and Implementation}

\subsection{Execution overview}

Section~\ref{sec:streaming_schedule} established that complete edge-weight
tensors are not required by the operator or its derivatives.  \sobek realizes
that lifetime property by ending edge-sized execution inside receiver
traversal: edge-local contractions reduce directly into bounded receiver-side
state, after which channel-space operations act on receiver rows.  Neither
$W$, an edge-message tensor, nor $\bar W$ crosses a kernel boundary as a global
tensor.

Here, \emph{receiver-factorized} means that work is organized around each
destination node rather than around a materialized edge tensor.  For every
receiver and output group, contributions from its incoming edges first reduce
into a bounded state $S_g$ that contains the angular, radial, and input-channel
dimensions but no output-channel dimension.  The learned output-channel mixing
is then applied once to that receiver state.  This separates graph traversal
from channel mixing and prevents either stage from producing a complete
edge-weight or edge-message array.

\begin{figure}[!htbp]
  \centering
  \makebox[\linewidth][c]{%
    \resizebox{\linewidth}{!}{\input{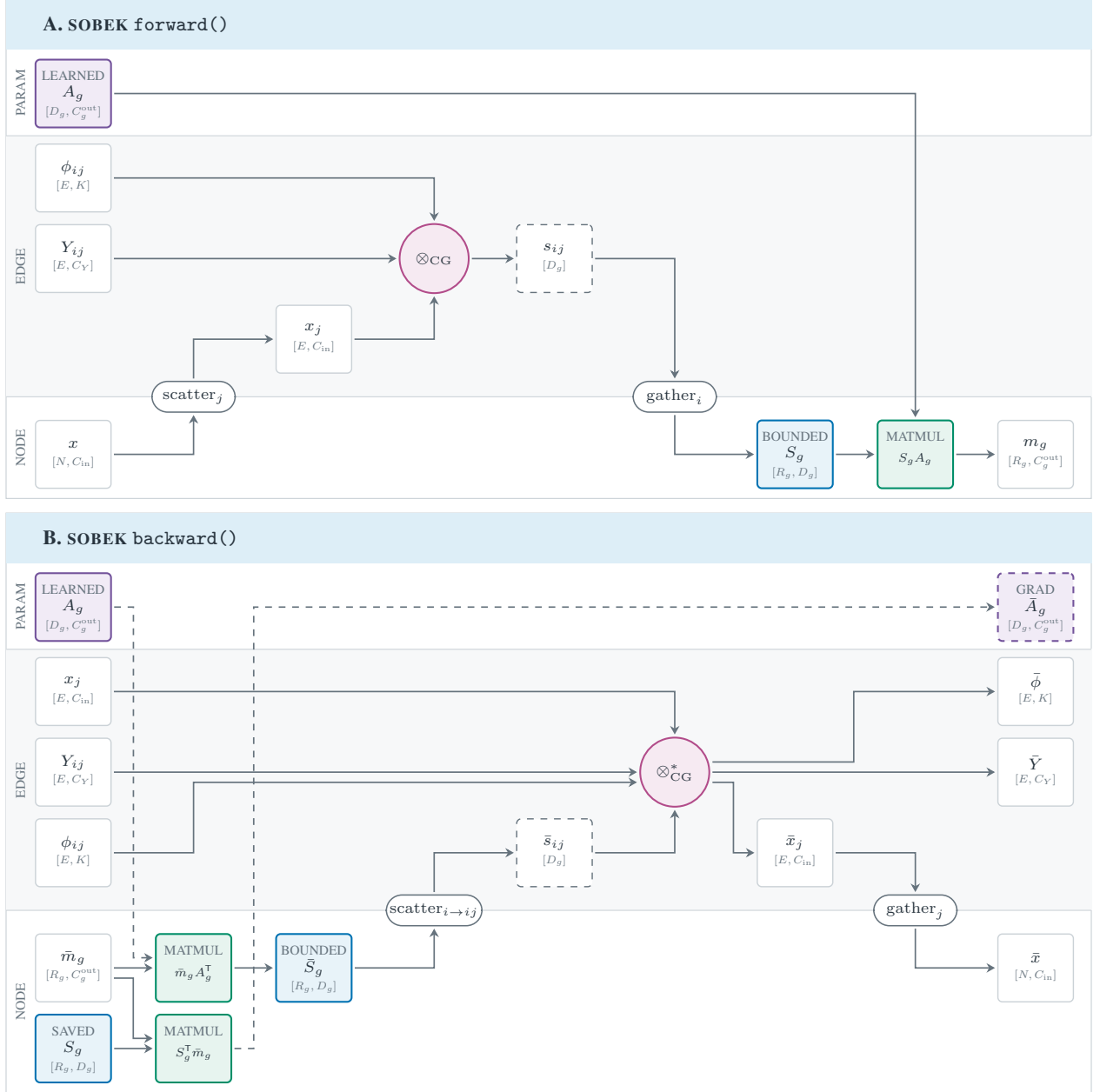}}%
  }
  \caption{Receiver-factorized \sobek dataflow for one output group.
  Lanes distinguish parameter-, edge-, and node-indexed values.  Purple, blue,
  green, and magenta mark learned parameters, bounded receiver state, channel
  matrix products, and edge-local CG operations, respectively; rounded
  scatter/gather nodes change graph ownership.  Dashed gray boxes are
  edge-local transient values, while the dashed purple box is the accumulated
  parameter gradient.  No complete edge-weight, edge-message, or
  edge-weight-adjoint tensor appears.}
  \label{fig:streaming_dataflow}
\end{figure}

Figure~\ref{fig:streaming_dataflow} summarizes the receiver-factorized forward
and backward dataflow for one output group $g$.  It is an algebraic view rather
than a kernel-launch trace.  Let $E=|\E|$ and $N=|\V|$; $R_g$ counts
receiver/output-component rows, and
$D_g=\sum_{\pi\in\mathcal P_g}K C^{\mathrm{in}}_\pi$ is the flattened path,
radial, and input-multiplicity width.

The edge-local CG map combines $\phi_{ij}$, $x_j$, and $Y_{ij}$ into
$s_{ij}$.  Receiver gather forms $S_g$, and channel mixing produces
$m_g=S_gA_g$.  Backward reverses this flow: channel mixing forms $\bar S_g$,
receiver scatter supplies edge-local adjoints, and deterministic source gather
performs the cross-edge reduction for $\bar x$.  The following subsections give
the corresponding contractions and generated execution.

\subsection{Receiver-factorized forward}
\label{sec:receiver_architecture}

\sobek realizes conceptual streaming by reassociating the contraction around
each receiver.  Let $g$ denote an output-irrep group and $\mathcal{P}_g$ the
paths that contribute to it.  For a path $\pi$, input multiplicity $u$, and
output-irrep component $m_o$, define the unweighted angular contraction

\begin{equation}
  C_{\pi,u,m_o}(x_j,Y_{ij})
  =
  \sum_{m_i,m_y}
  c^{\pi}_{m_i,m_y,m_o}
  x_{j,\pi,u,m_i}
  Y_{ij,\pi,m_y},
  \label{eq:angular_path_contraction}
\end{equation}

where $c^{\pi}_{m_i,m_y,m_o}$ contains the path normalization and sparse
Clebsch--Gordan coefficient.  A receiver-contiguous traversal accumulates

\begin{equation}
  S_{i,\pi,q,u,m_o}
  =
  \sum_{j\in\N(i)}
  \phi_{ij,q}
  C_{\pi,u,m_o}(x_j,Y_{ij}).
  \label{eq:receiver_state}
\end{equation}

For a fully connected \texttt{uvw} instruction, channel mixing then gives

\begin{equation}
  m^{(g)}_{i,w,m_o}
  =
  \sum_{\pi\in\mathcal{P}_g}
  \sum_{q=1}^{K}
  \sum_{u=1}^{C^{\mathrm{in}}_{\pi}}
  S_{i,\pi,q,u,m_o}
  A_{q,\pi,u,w}.
  \label{eq:receiver_mixing}
\end{equation}

For each output group, \sobek packs the receiver and output-irrep component
indices as rows, the radial and input-multiplicity indices as the contracted
dimension, and the output multiplicity as columns.
Equation~\ref{eq:receiver_mixing} is then the channel matrix product

\begin{equation}
  m = S A.
  \label{eq:receiver_matrix_product}
\end{equation}

Equations~\ref{eq:receiver_state}--\ref{eq:receiver_matrix_product} are an
exact reassociation of Eq.~\ref{eq:streaming_expression}.  The receiver state
has no output-multiplicity axis and does not scale with the number of edges.
A generated receiver traversal constructs $S$, a channel GEMM applies $A$,
and a compact write kernel places the result in the output tensor.  Forward
states needed for radial-parameter gradients are retained only
as complete output groups within the fixed scratch budget; other groups are
recomputed over bounded receiver ranges.  This policy changes reuse and
recomputation, not the streaming transformation itself.

\subsection{Receiver-factorized backward}
\label{sec:receiver_adjoint}

Backward differentiates the same receiver program.  With receiver and
output-irrep component indices packed as the row dimension of
Eq.~\ref{eq:receiver_matrix_product}, the two channel-space adjoints are

\begin{equation}
  \bar S = \bar m A^{\mathsf T},
  \qquad
  \bar A = S^{\mathsf T}\bar m.
  \label{eq:receiver_adjoint}
\end{equation}

Receiver/channel GEMMs form $\bar S$ and accumulate $\bar A$.  Generated
edge-owned kernels differentiate Eq.~\ref{eq:receiver_state} to form
$\bar\phi$ and $\bar Y$ in receiver-sorted edge order; these quantities have a
unique output location per edge and require no cross-edge reduction.  A
separate source-contiguous CSR traversal owns $\bar x$: one source row has one
fixed reduction order, avoiding both atomics and nondeterministic
accumulation.  High-work receiver recomputation uses the same grouped
output-component traversal as forward, while edge and source kernels assign
disjoint edge or channel ownership so that their reductions remain local.

Compatible $\bar S$ states share a source traversal when they fit the scratch
bound; otherwise \sobek chunks output groups and receiver ranges while keeping
retained receiver-side state under the same fixed budget.  These choices change
reuse and recomputation, not the adjoint equations or reduction ownership.
The detailed fit-aware source policy is given in
\ref{app:source_schedule}.

Crucially, backward does not form an edge-weight adjoint $\bar W$ and then
consume it incrementally.  Radial, angular, and node-feature gradients are
obtained directly from $S$, $\bar S$, and the graph traversals, so $\bar W$ is
never constructed, not even as an edge chunk.

\subsection{Double backward as the differentiated receiver program}

Double backward is differentiation of
Eqs.~\ref{eq:receiver_state}--\ref{eq:receiver_adjoint}, rather than an
unrelated collection of higher-order kernels.  A receiver traversal constructs
the primal state $S$ and its active directional state $dS$.  The corresponding
channel identities are

\begin{align}
  dm &= dS\,A + S\,dA, \\
  d\bar S &= d\bar m\,A^{\mathsf T}
             + \bar m\,dA^{\mathsf T}, \\
  d\bar A &= dS^{\mathsf T}\bar m
             + S^{\mathsf T}d\bar m.
  \label{eq:receiver_double_adjoint}
\end{align}

GEMMs construct output tangents, adjoints, and radial-parameter derivatives
from the same bounded receiver states.  Forward-retained $S$ is reused, while
the additional primal, adjoint, and directional states follow the same
fit-and-recompute policy as first backward.  Receiver CSR owns edge-local
Hessian terms for $\phi$ and $Y$, and source CSR owns deterministic node
adjoints.  Thus the receiver, edge, and source ownership extends through second
order without a separate kernel architecture or edge-scaled state;
\ref{app:source_schedule} summarizes the shared scratch rule.

This organization builds directly on the higher-order decomposition developed
by the OpenEquivariance authors~\cite{bharadwaj2025openequivariance}, expressing double backward as forward-style and
backward-style replay terms and lowers those terms into dedicated receiver- and
source-major kernels.  \sobek lowers the same mathematical decomposition into
the receiver program above.

\subsection{Generated deterministic runtime and execution contracts}

\sobek accepts input, output, and spherical-harmonic irrep structures and
generates specialized kernels for forward, backward, and double backward.

A thin Python/PyTorch wrapper validates the contract, adapts layout, prepares
the graph, and connects generated entry points to autograd.  Following
OpenEquivariance~\cite{bharadwaj2025openequivariance}, \sobek treats the static
tensor-product description as input to a kernel generator.  It enumerates
valid paths and sparse Clebsch--Gordan terms and emits CUDA source code with compile-time paths, loop bounds, offsets, and coefficients.
PyTorch compiles and caches the extension, so steady-state execution contains
neither an instruction interpreter nor code-generation overhead.

\sobek compiles channelwise \texttt{uvu} directly rather than pruning
off-diagonal mixing from dense \texttt{uvw}.  Its deterministic forward,
backward, and double-backward schedule preserves tied-channel semantics without
edge workspace.

Generated kernels use \texttt{ir\_mul} layout, with adjacent multiplicity
channels contiguous within each irrep block.  The wrapper transposes the default
\texttt{mul\_ir} layout at the boundary, or callers can retain \texttt{ir\_mul}
across layers.  Cached stable CSR views give receiver and source reductions fixed
ownership without nondeterministic atomics.  CUDA Graph capture and TF32 channel
GEMMs are optional; all reported comparisons use eager fp32 with TF32 disabled.
\ref{app:additional_implementation} provides additional CUDA
implementation details, including compile-time specialization, kernel grouping,
and runtime scratch-budget policy.

\section{Evaluation}
\label{sec:evaluation}

Our evaluation asks three questions:
\begin{enumerate}
  \setlength{\topsep}{3pt}
  \setlength{\itemsep}{1pt}
  \setlength{\parsep}{0pt}
  \setlength{\parskip}{0pt}
  \item \textbf{Memory scaling.} How does peak memory scale with edge count and
    operator width?
  \item \textbf{Throughput scaling.} Does throughput improve as larger workloads
    fill the GPU under both degree and graph-count scaling?
  \item \textbf{Generality across feature structures.} Does the approach preserve its
    correctness, memory, and throughput advantages across varied node-feature
    representations, including fully connected \texttt{uvw} and channelwise
    \texttt{uvu}?
\end{enumerate}
Forward, full first backward, and native double backward are included in every
sweep and comparison rather than treating differentiation order as a separate
question.

Answering all three questions requires a common reference for scaling,
throughput, and numerical equivalence.  Native e3nn~\cite{geiger2022e3nn} is a
general-purpose reference, while cuEquivariance~\cite{nvidia2025cuequivariance}
also provides a fused fully connected tensor-product convolution with graph
gather/scatter.  We choose
OpenEquivariance~\cite{bharadwaj2025openequivariance} as the optimized reference
because it provides the directly matched deterministic forward, backward, and
double-backward contract required by our comparison and validation protocol.

\subsection{Reference and experimental design}

\paragraph{Comparison and validation policy.}
We compare both backends on an NVIDIA RTX 5090 using fp32 throughout.  Every
presented pair receives the same complete input: graph topology and ordering,
features, weights, and all other tensors.  All \sobek measurements use a 128 MiB
receiver scratch budget and highest-precision fp32 channel GEMMs.  Compilation,
graph generation, and graph preparation are outside timed execution.

Every data point presented in this paper passes strict double-peer validation:
two \sobek evaluations match exactly, two OpenEquivariance evaluations match
exactly, and all four agree within the cross-backend tolerance.  The base
tolerance is $\mathrm{rtol}=\mathrm{atol}=5\times10^{-4}$; large or dense
reductions use $\mathrm{rtol}=2\times10^{-3}$ and
$\mathrm{atol}=2\times10^{-2}$.  These tolerances accommodate changes in the
order of floating-point additions between the two schedules.

Each scaling point runs in a fresh process.  After at least five calls and
50 ms of warm-up, CUDA-event timing collects 10--30 samples totaling at least
250 ms and reports the median.  Peak allocated CUDA memory is measured after
compilation warm-up in the same isolated worker.  A confirmed OOM is recorded
at the attempted graph size, and no later point is fabricated for that backend.

\paragraph{Scaling axes.}
For \emph{degree scaling}, we construct radius graphs on one set of positions
for 2,048 nodes, varying the cutoff to obtain mean receiver degrees from 4 to
1,023.  For \emph{graph-count scaling}, we repeat one 128-node, mean-degree-32
graph from $B=2$ to 511, increasing total $N$ from 256 to 65,408.  Both axes
span $8.2\times10^3$ to $2.1\times10^6$ edges, separating longer neighborhoods
from proportional node-and-edge scaling.  Both backends are measured until OOM;
Appendix Table~\ref{tab:graph_scaling_contracts} gives the exact grids and
realized degree statistics.

\paragraph{Representative irrep profiles.}

Input/output irreps are equal with alternating even/odd parity; spherical
harmonics contain one copy per order through $L_{\max}$, and the edge embedding
has width $K=16$.  The
$L_{\max}=1,2,3$ profiles are $128/64$, $128/64/64$, and $128/64/64/64$.
Holding 128 scalar channels and 64 channels in every included non-scalar irrep
ensures that angular arithmetic and radial-weight count rise together:
$N_{\mathrm w}=36{,}864$, $73{,}728$, and $131{,}072$.

For fp32 materialization, $M_W=4|\E|N_{\mathrm w}$ bytes.  Thus $W$ alone
reaches 32 GiB at approximately $2.3\times10^5$, $1.2\times10^5$, and
$6.6\times10^4$ directed edges for $128/64$, $128/64/64$, and
$128/64/64/64$, respectively.  Inputs, outputs, library state, and temporary
storage make the practical capacity lower.

\subsection{Memory scaling}

\begin{figure}[!htbp]
  \centering
  \includegraphics[width=\linewidth]{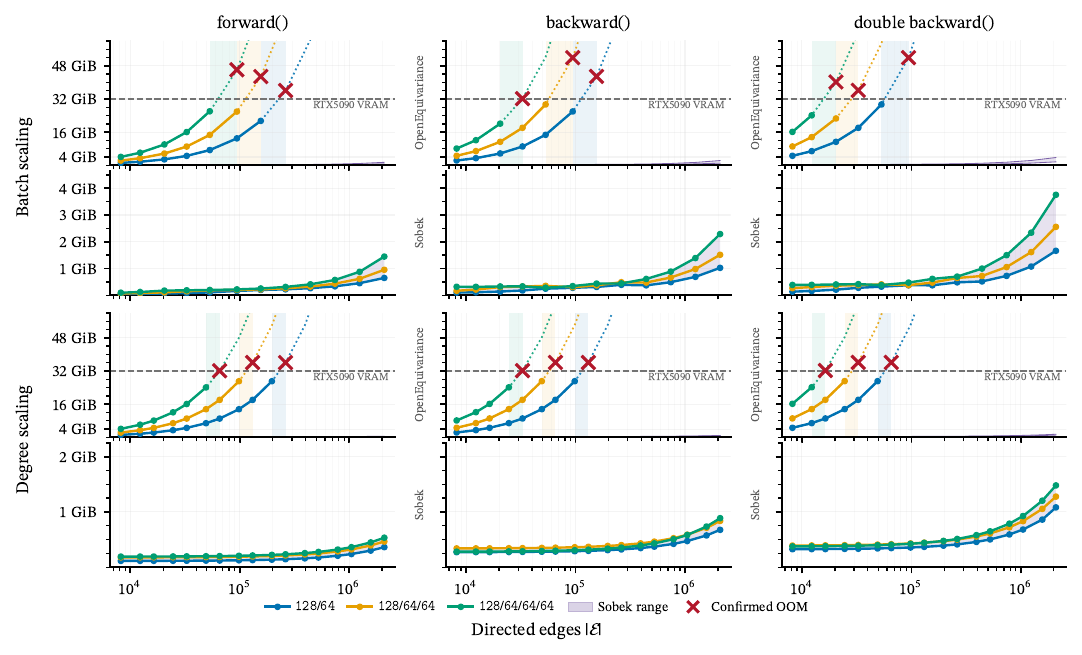}
  \caption{Peak allocated CUDA memory under graph-count (top) and degree
  (bottom) scaling, with differentiation order on columns.  Each panel separates
  OpenEquivariance and \sobek linear ranges.  Profile colors identify
  multiplicity schedules; purple bands show \sobek's cross-profile range and
  repeat it on the OpenEquivariance axes for scale.  Dotted curves show $M_W$
  and the dashed line marks 32 GiB RTX~5090 capacity; shaded intervals end at
  confirmed OOMs (red crosses).}
  \label{fig:density_memory}
\end{figure}

We first seek the practical memory scaling law: whether peak allocation follows
the materialized $M_W\propto|\E|N_{\mathrm w}$ term as edge count and operator
width increase, or whether \sobek replaces that term with bounded workspace.
Figure~\ref{fig:density_memory} compares peak allocated memory under both
scaling constructions and all three differentiation orders.  At the last degree
point shared by both backends, OpenEquivariance allocates 24.1--27.1 GiB across
the nine profile/pass pairs, while \sobek allocates 0.1--0.4 GiB.
OpenEquivariance's confirmed OOMs occur between $1.6\times10^4$ and
$2.6\times10^5$ edges, depending on profile and differentiation order.  Every
\sobek curve reaches the campaign endpoint of $2.1\times10^6$ edges, where peak
allocation remains 0.4--1.5 GiB.

The result is not constant total memory.  \sobek removes the dominant
$O(|\E|N_{\mathrm w})$ allocation by never materializing per-edge radial
weights or edge messages, and caps tensor-product scratch at 128 MiB.  The
graph itself still requires edge indices, spherical harmonics, embeddings, and
their saved gradients, all of which grow with $|\E|$; under graph-count scaling,
node features, outputs, and node-scaled intermediates grow with $N$ as well.
Consequently, fixed-$N$ degree scaling retains a much smaller residual
edge-linear component, while graph-count scaling combines that component with
node-linear storage.  Operator width changes these constants, but no longer
multiplies edge count by the full radial-weight count.

\FloatBarrier
The observed law is therefore bounded tensor-product workspace plus unavoidable
graph and feature storage, rather than $O(1)$ memory overall.

\subsection{Throughput scaling}

\begin{figure}[!htbp]
  \centering
  \includegraphics[width=\linewidth]{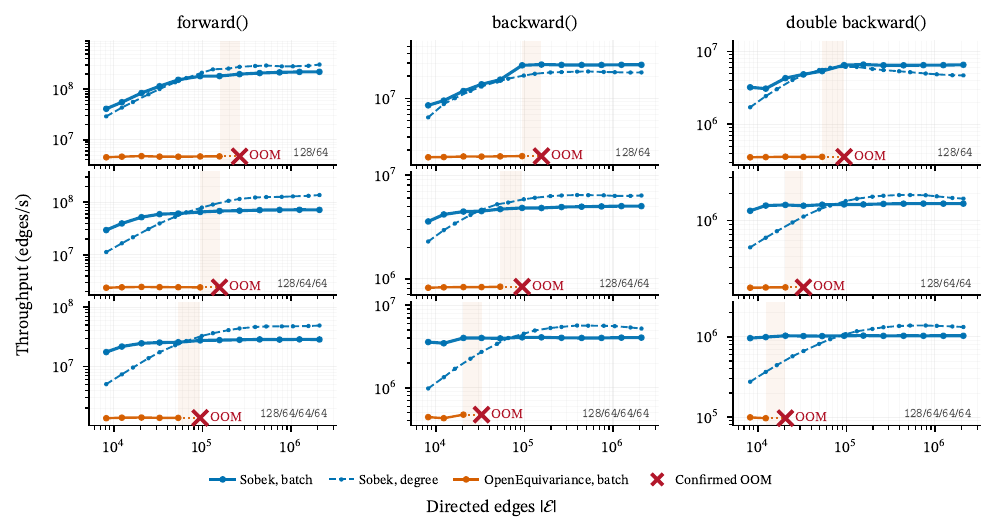}
  \caption{Absolute throughput by profile row, with differentiation order on columns.
  Solid and dashed blue curves show \sobek graph-count and degree scaling;
  orange shows OpenEquivariance graph-count scaling.  Both axes are logarithmic,
  with panel-specific throughput ranges.  OpenEquivariance degree curves are
  omitted because they overlap graph-count curves.  Orange bands and dotted
  segments connect the last success to confirmed OOMs (red crosses).}
  \label{fig:density_throughput}
\end{figure}

We next ask whether removing edge-weight materialization preserves processing
rate as well as capacity.  Graph-count scaling measures how throughput changes
as larger workloads fill the GPU, while degree scaling tests whether longer
receiver neighborhoods change per-edge execution efficiency.
Figure~\ref{fig:density_throughput} shows the scaling law directly.
OpenEquivariance throughput is effectively constant with workload for each
profile and pass; its degree curves overlap its graph-count curves and are
therefore omitted from the figure.  Every point completed by both backends was
nevertheless measured and passed strict double-peer validation.  At the last
degree workload shared by both backends, \sobek speedups span
$3.8$--$54.4\times$ across the nine profile/pass pairs.  At the exact
OpenEquivariance graph-count capacity, speedups across the three primary
profiles are $19.7$--$42.1\times$ for forward,
$5.6$--$16.9\times$ for backward, and $7.5$--$15.0\times$ for double
backward.

All \sobek graph-count curves reach $2.1\times10^6$ edges and retain at least
99.2\% of their peak throughput at that endpoint.  Under degree scaling,
forward peaks at the endpoint, backward retains 91.8--99.0\% of peak, and
double backward retains 76.3--95.9\%.  For the three primary profiles,
extending graph-count scaling from the OpenEquivariance capacity to $B=511$
increases work per call by $9.5$--$170.3\times$; \sobek's maximum throughput
over the extended curve is $1.0$--$1.2\times$ its throughput at the matched
capacity.  The larger feasible workloads therefore primarily increase usable
problem size while maintaining or modestly improving throughput.

This behavior follows directly from the avoided memory operations.  A
materialized \texttt{uvw} schedule computes $N_{\mathrm w}$ radial weights per
edge, writes them to GPU memory, and reads them back for the contraction;
$N_{\mathrm w}$ grows quadratically with multiplicity.  \sobek consumes these
values without copying an $|\E|\times N_{\mathrm w}$ tensor through global
memory.

\FloatBarrier
Removing those single-use writes and reads avoids OpenEquivariance's early
capacity boundary and permits larger batches without increasing the bounded
tensor-product workspace.

\subsection{Generality across feature structures}
\label{sec:generality}

We use capacity matching to compare varied representations fairly.  For each
profile, we independently determine for each pass the maximum batch size (graph
count) $B_{\mathrm{oeq}}$ successfully completed by OpenEquivariance, then time
both backends on exactly the same input and double-peer cross-validate their
outputs to ensure the same numerical result under the strict policy above.
This gives the largest workload that both OpenEquivariance and \sobek can
execute.  These are
OpenEquivariance-capacity-limited common workloads, not \sobek throughput
ceilings.  Capacity matching therefore understates both usable problem size and
throughput at larger, better-occupying batches.

The main campaign covers sixteen fully connected \texttt{uvw} profiles on the
fixed graph.  The grid varies angular order, channel width, scalar fraction,
and radial-operator size.  It tests whether \sobek's gains persist as feature
structure shifts from scalar-heavy to more evenly distributed representations,
as higher-order irreps add angular arithmetic, and as wider operators increase
channel mixing and radial-weight count.  This breadth establishes generality
beyond the three profiles chosen to expose the scaling laws.

\begin{table}[!htbp]
  \centering
  \caption{Capacity-matched \texttt{uvw} results on the fixed 128-node, degree-32 graph. Each pass uses OpenEquivariance's largest graph count $B_{\mathrm{oeq}}$. Bold profiles are those used in the scaling-law figures. Paired Peak GiB and M edges/s cells use S $|$ oeq order. Peak GiB is peak allocated CUDA memory. Profiles list multiplicities; $N_{\mathrm w}$ is the radial-weight count.}
  \label{tab:capacity_matched}
  \scriptsize
  \setlength{\tabcolsep}{2.7pt}
  \renewcommand{\arraystretch}{1.08}
  \resizebox{\linewidth}{!}{%
  \begin{tabular}{@{}lrrrrrrrrrrrrr@{}}
    \toprule
    Profile & $N_{\mathrm w}$ & \multicolumn{4}{c}{\texttt{forward()}} & \multicolumn{4}{c}{\texttt{backward()}} & \multicolumn{4}{c}{\texttt{double backward()}} \\
    \cmidrule(lr){3-6}\cmidrule(lr){7-10}\cmidrule(lr){11-14}
    & & Graphs & Peak GiB & M edges/s & Ratio & Graphs & Peak GiB & M edges/s & Ratio & Graphs & Peak GiB & M edges/s & Ratio \\
    & & $B_{\mathrm{oeq}}$ & S $|$ oeq & S $|$ oeq & S/oeq & $B_{\mathrm{oeq}}$ & S $|$ oeq & S $|$ oeq & S/oeq & $B_{\mathrm{oeq}}$ & S $|$ oeq & S $|$ oeq & S/oeq \\
    \midrule
    $32/16$ & 2,304 & 862 & 0.7 | 30.7 & 661.0 | 79.9 & \cellcolor{sobekblue!30}\textbf{8.28$\times$} & 431 & 0.6 | 30.7 & 104.7 | 24.4 & \cellcolor{sobekblue!22}\textbf{4.29$\times$} & 215 & 0.7 | 30.6 & 24.3 | 6.1 & \cellcolor{sobekblue!22}\textbf{4.00$\times$} \\
    $32/32$ & 4,096 & 487 & 0.5 | 30.7 & 723.5 | 45.7 & \cellcolor{sobekblue!30}\textbf{15.84$\times$} & 243 & 0.6 | 30.7 & 91.9 | 11.7 & \cellcolor{sobekblue!22}\textbf{7.88$\times$} & 121 & 0.5 | 30.5 & 22.9 | 3.5 & \cellcolor{sobekblue!22}\textbf{6.54$\times$} \\
    $32/16/16$ & 4,608 & 433 & 0.5 | 30.8 & 188.3 | 41.7 & \cellcolor{sobekblue!22}\textbf{4.52$\times$} & 216 & 0.6 | 30.7 & 17.4 | 12.7 & \cellcolor{sobekblue!8}\textbf{1.37$\times$} & 108 & 0.5 | 30.6 & 5.3 | 3.1 & \cellcolor{sobekblue!8}\textbf{1.73$\times$} \\
    $32/32/32$ & 11,264 & 178 & 0.3 | 30.8 & 170.4 | 16.0 & \cellcolor{sobekblue!30}\textbf{10.62$\times$} & 88 & 0.4 | 30.4 & 11.9 | 5.1 & \cellcolor{sobekblue!14}\textbf{2.32$\times$} & 44 & 0.4 | 30.4 & 3.7 | 1.2 & \cellcolor{sobekblue!14}\textbf{3.00$\times$} \\
    $32/16/16/16$ & 8,192 & 244 & 0.4 | 30.7 & 85.9 | 23.3 & \cellcolor{sobekblue!14}\textbf{3.68$\times$} & 121 & 0.5 | 30.5 & 12.7 | 6.7 & \cellcolor{sobekblue!8}\textbf{1.89$\times$} & 60 & 0.4 | 30.2 & 3.3 | 1.7 & \cellcolor{sobekblue!8}\textbf{1.93$\times$} \\
    $32/32/32/32$ & 23,552 & 85 & 0.3 | 30.7 & 70.8 | 8.3 & \cellcolor{sobekblue!30}\textbf{8.48$\times$} & 42 & 0.3 | 30.3 & 6.9 | 2.7 & \cellcolor{sobekblue!14}\textbf{2.54$\times$} & 21 & 0.4 | 30.3 & 1.8 | 0.6 & \cellcolor{sobekblue!14}\textbf{2.99$\times$} \\
    \specialrule{0.25pt}{2pt}{2pt}
    $64/32$ & 9,216 & 217 & 0.3 | 30.7 & 423.9 | 20.9 & \cellcolor{sobekblue!38}\textbf{20.28$\times$} & 108 & 0.4 | 30.5 & 58.8 | 6.4 & \cellcolor{sobekblue!30}\textbf{9.14$\times$} & 54 & 0.4 | 30.5 & 14.3 | 1.6 & \cellcolor{sobekblue!30}\textbf{9.16$\times$} \\
    $64/64$ & 16,384 & 122 & 0.3 | 30.6 & 345.7 | 12.2 & \cellcolor{sobekblue!38}\textbf{28.31$\times$} & 61 & 0.3 | 30.6 & 43.3 | 3.7 & \cellcolor{sobekblue!30}\textbf{11.59$\times$} & 30 & 0.4 | 30.1 & 9.2 | 0.8 & \cellcolor{sobekblue!30}\textbf{10.83$\times$} \\
    $64/32/32$ & 18,432 & 108 & 0.3 | 30.5 & 125.7 | 10.7 & \cellcolor{sobekblue!30}\textbf{11.79$\times$} & 54 & 0.3 | 30.5 & 9.8 | 3.2 & \cellcolor{sobekblue!14}\textbf{3.03$\times$} & 27 & 0.4 | 30.5 & 3.1 | 0.7 & \cellcolor{sobekblue!22}\textbf{4.25$\times$} \\
    $64/64/64$ & 45,056 & 44 & 0.2 | 30.3 & 86.4 | 4.3 & \cellcolor{sobekblue!38}\textbf{19.94$\times$} & 22 & 0.3 | 30.4 & 5.5 | 1.4 & \cellcolor{sobekblue!22}\textbf{4.07$\times$} & 11 & 0.4 | 30.4 & 1.7 | 0.3 & \cellcolor{sobekblue!22}\textbf{5.82$\times$} \\
    $64/32/32/32$ & 32,768 & 61 & 0.2 | 30.6 & 59.7 | 5.3 & \cellcolor{sobekblue!30}\textbf{11.28$\times$} & 30 & 0.3 | 30.1 & 7.9 | 1.9 & \cellcolor{sobekblue!22}\textbf{4.20$\times$} & 15 & 0.4 | 30.1 & 2.0 | 0.4 & \cellcolor{sobekblue!22}\textbf{5.05$\times$} \\
    $64/64/64/64$ & 94,208 & 21 & 0.2 | 30.3 & 34.6 | 1.9 & \cellcolor{sobekblue!38}\textbf{18.04$\times$} & 10 & 0.3 | 28.9 & 3.2 | 0.7 & \cellcolor{sobekblue!22}\textbf{4.93$\times$} & 5 & 0.4 | 28.9 & 0.8 | 0.1 & \cellcolor{sobekblue!22}\textbf{5.79$\times$} \\
    \specialrule{0.25pt}{2pt}{2pt}
    \textbf{128/64} & 36,864 & 54 & 0.2 | 30.5 & 197.0 | 4.7 & \cellcolor{sobekblue!38}\textbf{42.13$\times$} & 27 & 0.3 | 30.5 & 28.1 | 1.7 & \cellcolor{sobekblue!38}\textbf{16.95$\times$} & 13 & 0.3 | 29.4 & 5.4 | 0.4 & \cellcolor{sobekblue!30}\textbf{14.99$\times$} \\
    $128/128$ & 65,536 & 30 & 0.2 | 30.1 & 132.2 | 2.7 & \cellcolor{sobekblue!38}\textbf{49.68$\times$} & 15 & 0.3 | 30.1 & 14.0 | 1.0 & \cellcolor{sobekblue!30}\textbf{14.73$\times$} & 7 & 0.3 | 28.1 & 4.0 | 0.2 & \cellcolor{sobekblue!38}\textbf{19.87$\times$} \\
    \textbf{128/64/64} & 73,728 & 27 & 0.2 | 30.4 & 66.8 | 2.5 & \cellcolor{sobekblue!38}\textbf{27.21$\times$} & 13 & 0.3 | 29.4 & 4.7 | 0.8 & \cellcolor{sobekblue!22}\textbf{5.60$\times$} & 6 & 0.4 | 27.1 & 1.4 | 0.2 & \cellcolor{sobekblue!22}\textbf{7.51$\times$} \\
    \textbf{128/64/64/64} & 131,072 & 15 & 0.2 | 30.1 & 26.8 | 1.4 & \cellcolor{sobekblue!38}\textbf{19.71$\times$} & 7 & 0.3 | 28.1 & 4.0 | 0.5 & \cellcolor{sobekblue!30}\textbf{8.46$\times$} & 3 & 0.4 | 24.1 & 1.0 | 0.1 & \cellcolor{sobekblue!30}\textbf{10.24$\times$} \\
    \bottomrule
  \end{tabular}
  }
\end{table}

Across the grid, \sobek is faster in all 48 profile/pass comparisons, with
geometric-mean speedups of $14.9\times$, $5.0\times$, and $5.7\times$ for
forward, backward, and double backward.  Individual comparisons range from
$1.4\times$ to $49.7\times$.  Its matched peak allocation is
0.2--0.7 GiB versus 24.1--30.8 GiB, and every point uses zero
edge-weight workspace.

We repeat the same protocol for homogeneous channelwise \texttt{uvu}, using
three equal-multiplicity widths at each angular order.  \sobek is faster in all
27 profile/pass comparisons.  Geometric-mean speedups are $3.3\times$,
$3.4\times$, and $1.8\times$ for forward, backward, and double backward, with
respective ranges of $1.2$--$5.9\times$, $2.5$--$4.2\times$, and
$1.6$--$2.1\times$.  Matched peak allocation is 1.7--4.5 GiB for \sobek versus
30.3--30.8 GiB for OpenEquivariance, and every \sobek point uses zero
edge-weight workspace.  Appendix Table~\ref{tab:uvu_capacity_matched} reports
the complete grid.

The smaller \texttt{uvu} speedups are consistent with the same mechanism.
Tied-channel mixing removes an output-multiplicity factor from the materialized
edge-weight expansion, so streaming eliminates a smaller fraction of total
work; higher angular order also raises the relative cost of Clebsch--Gordan
contractions.  The timing and peak-allocation results support this
explanation based on avoided memory reads and writes but do not isolate it from
occupancy, cache locality, launch overhead, or kernel mapping.

\section{Related Work}

\paragraph{Memory-efficient exact operators and streaming reformulations.}
Two close systems analogues are KeOps and FlashAttention.  KeOps generates GPU
map-reductions over symbolic pairwise arrays, including automatic
differentiation, without materializing the complete kernel or distance
matrix~\cite{charlier2021keops}.  FlashAttention preserves exact attention
while changing its evaluation order so that the dominant global intermediate
is never materialized~\cite{dao2022flashattention}.
FlashAttention-2 further shows that work partitioning and hardware mapping
remain important after the algorithmic memory
transformation~\cite{dao2023flashattention2}.  \sobek applies the same separation between a
streaming formulation and its hardware realization to irregular equivariant
graph convolution: streaming determines which intermediates survive globally,
while receiver factorization and generated CUDA determine how the resulting
computation maps to the GPU.

\paragraph{Equivariant kernels and generated GPU systems.}
OpenEquivariance generates sparse, problem-specialized kernels for equivariant
tensor products and deterministic graph
convolutions~\cite{bharadwaj2025openequivariance}.  NVIDIA cuEquivariance
provides optimized CUDA execution for a general segmented-tensor-product
representation~\cite{nvidia2025cuequivariance}, and recent work has studied its
integration and reduced-precision execution in
MACE~\cite{benoit2025speedingmace}.  FlashTP combines kernel fusion,
Clebsch--Gordan sparsity, and path aggregation to reduce intermediate memory
traffic in equivariant interatomic potentials~\cite{lee2025flashtp}.  \sobek shares their use of static
tensor-product descriptions and specialized kernels but places its main
optimization boundary around the complete edge-conditioned convolution, where
the lifetime of radial weights and their adjoints determines peak memory.

\paragraph{Alternative operators and connection modes.}
Other work reduces cost by changing the equivariant operator or its channel
connectivity.  eSCN reduces $\mathrm{SO}(3)$ convolutions to operations in
$\mathrm{SO}(2)$, and EquiformerV2 uses this construction for higher-degree
representations~\cite{passaro2023escn,liao2024equiformerv2}.  Equiformer and
InfGCN instead use depthwise or channelwise tensor products to avoid fully
connected multiplicity mixing within the tensor
product~\cite{liao2023equiformer,cheng2023infgcn}.
Xie et al.~\cite{xie2025price} formalize the resulting distinction between
arithmetic scaling and the equivariant bilinear maps an operator can represent.
These choices can reduce arithmetic or parameter cost, but they do not by
themselves eliminate replication over graph edges in a materialized schedule.
\sobek is orthogonal: it provides native \texttt{uvu} execution for tied-channel
models while its streamed \texttt{uvw} path retains fully connected couplings
without the $|\E|$-scaled weight intermediate.

\section{Conclusion}

The central result of this work is that the large global intermediates commonly
associated with edge-conditioned equivariant convolutions are consequences of
their evaluation schedule, not requirements of the operator.  Forward, first
backward, and double backward can instead be organized as streamed edge-local
products consumed immediately by receiver-, source-, or edge-owned reductions.
This reformulation preserves fully connected multiplicity mixing and the exact
derivative structure while replacing edge-scaled workspace with bounded
receiver-side state and recomputation.

\sobek realizes this reformulation as a generated-CUDA backend.  Its
receiver-factorized schedules evaluate radial-to-CG projections without
materializing the edge-wise weight and message tensors that dominate memory in
dynamic molecular graphs.  The implementation supports forward, first
backward, and double backward with bounded auxiliary state and without
constructing the edge-weight adjoint.

Extending the same bounded-lifetime principle to eSCN-style edge-aligned
$\mathrm{SO}(2)$ kernels is a particularly promising direction, combining
their lower angular complexity with streaming execution at higher angular
resolution~\cite{passaro2023escn}.

By separating operator expressivity from edge-sized workspace, streaming
execution relaxes the tradeoffs among neighborhood size, representation width,
batch size, and channel connectivity.  This could make it practical to run
full-atom machine-learning interatomic potentials on proteins and other large
biomolecular assemblies, use longer-range or higher-resolution representations
without collapsing batch size, and train or sample large equivariant generative
models over substantially larger structures.  More broadly, it would let model
designers choose representation width and connectivity for physical fidelity
rather than for the peak memory of a transient edge tensor.  We hope this
systems result translates into broader capabilities throughout the geometric
deep learning and molecular modeling ecosystem.

\clearpage
\appendix
\renewcommand{\thesection}{Appendix \Alph{section}}

\section{Benchmark Regimes}
\label{app:benchmark_regimes}

\begin{table}[!htbp]
  \centering
  \caption{Degree and graph-count parameter grids.  Degree statistics are realized
  over receiver nodes, with mean $\bar d$ and standard deviation $\sigma_d$;
  $N$ is exact and $|\E|$ is rounded to two significant figures.}
  \label{tab:graph_scaling_contracts}
  \scriptsize
  \setlength{\tabcolsep}{2.7pt}
  \renewcommand{\arraystretch}{1.08}
  \begin{minipage}[t]{0.58\linewidth}
    \centering
    \textbf{(a) Degree: vary neighborhood size}\par
    {\itshape Fixed: $B=1$}\par
    {\itshape Nodes/graph $=$ total $N=2{,}048$}\par\smallskip
    \begin{tabular}{@{}rrr@{\qquad}rrr@{}}
      \toprule
      $\bar d$ & $\sigma_d$ & $|\E|$ &
      $\bar d$ & $\sigma_d$ & $|\E|$ \\
      \midrule
      4  & 1.95  & \scinum{8.2}{3} & 96    & 26.81  & \scinum{2.0}{5} \\
      6  & 2.50  & \scinum{1.2}{4} & 128   & 36.47  & \scinum{2.6}{5} \\
      8  & 3.01  & \scinum{1.6}{4} & 192   & 56.28  & \scinum{3.9}{5} \\
      12 & 3.89  & \scinum{2.5}{4} & 256   & 76.72  & \scinum{5.2}{5} \\
      16 & 4.76  & \scinum{3.3}{4} & 384   & 117.76 & \scinum{7.9}{5} \\
      24 & 6.61  & \scinum{4.9}{4} & 512   & 157.90 & \scinum{1.0}{6} \\
      32 & 8.59  & \scinum{6.6}{4} & 768   & 226.61 & \scinum{1.6}{6} \\
      48 & 12.83 & \scinum{9.8}{4} & 1,023 & 273.19 & \scinum{2.1}{6} \\
      64 & 17.30 & \scinum{1.3}{5} &       &        &                 \\
      \bottomrule
    \end{tabular}
  \end{minipage}%
  \hfill
  \begin{minipage}[t]{0.36\linewidth}
    \centering
    \textbf{(b) Graph count: vary $B$}\par
    {\itshape Fixed: nodes/graph $=128$}\par
    {\itshape $\bar d=32$; $\sigma_d=10.49$}\par\smallskip
    \begin{tabular}{@{}rrr@{}}
      \toprule
      $B$ & Total $N$ & $|\E|$ \\
      \midrule
      2   & 256    & \scinum{8.2}{3} \\
      3   & 384    & \scinum{1.2}{4} \\
      5   & 640    & \scinum{2.0}{4} \\
      8   & 1,024  & \scinum{3.3}{4} \\
      13  & 1,664  & \scinum{5.3}{4} \\
      23  & 2,944  & \scinum{9.4}{4} \\
      38  & 4,864  & \scinum{1.6}{5} \\
      64  & 8,192  & \scinum{2.6}{5} \\
      108 & 13,824 & \scinum{4.4}{5} \\
      181 & 23,168 & \scinum{7.4}{5} \\
      304 & 38,912 & \scinum{1.2}{6} \\
      511 & 65,408 & \scinum{2.1}{6} \\
      \bottomrule
    \end{tabular}
  \end{minipage}
\end{table}
\FloatBarrier

\section{Additional Implementation Details}
\label{app:additional_implementation}

\subsection{Static compile-time decisions}

For a fixed irrep/path contract, radial width, derivative signature, GEMM mode,
and scratch bound, the generated translation unit is invariant to graph shape.
Node and edge counts, degree distribution, CSR ordering, retained state, and
chunk schedule select only launches already present in the cached extension.

The remaining CUDA choices are resolved before source emission.  Radial inputs
use aligned \texttt{float4} loads when their width is divisible by four.  If a
\texttt{uvw} contract contains an output group of dimension at least five,
groups of dimension at least three are split into kernels covering at most
four output components.  Backward source paths use a statically selected warp,
8-lane subwarp, or cooperative CTA executor; the CTA form is emitted only when
its shared tile is at most 24~KiB.  Edge derivatives use either one 32-lane
owner or two 16-lane owners.  Native \texttt{uvu} double backward additionally
pairs compatible source paths and consecutive parameter-gradient groups; at
component dimension seven and above, singleton directional-forward kernels
contain at most four paths.  These decisions fix path grouping and launch
geometry without graph-dependent kernel dispatch.

\subsection{Runtime scratch-budget policy}
\label{app:source_schedule}

The configured budget $B$---128~MiB in all reported experiments---bounds
floating receiver-side state, not graph indices, inputs, or outputs.  In
\texttt{uvw} forward, a group with
component dimension $d_g$, receiver-state width $D_g$, and output multiplicity
$C_g$ processes at most
\begin{equation}
  n_g = \min\!\left(
    |\V|,
    \max\!\left(1,
      \left\lfloor\frac{B}{4d_g(D_g+C_g)}\right\rfloor
    \right)
  \right)
  \label{eq:receiver_chunk_budget}
\end{equation}
receivers at once for fp32 budget $B$.  When radial-parameter gradients are
needed, \texttt{uvw} retains the fitting subset of complete groups that
maximizes avoided sparse-CG recomputation, breaking ties by byte cost;
\texttt{uvu} retains all groups only when their combined state fits.  Other
groups are recomputed over bounded receiver chunks.

On first-backward overflow, contracts with at least twenty paths may share one
all-group node chunk; other contracts chunk groups independently.  A chunked
route is used only when it produces at least three chunks.  Its source-CSR views
and exact widths are prepared with the topology and reused by double backward.
The second-order planner derives required state from active cotangents, greedily
packs contiguous groups while retained state plus the largest temporary fits
$B$, then chunks receivers.  Forward-retained primal state is reused only when
it enables one all-group, all-receiver chunk.  Receiver and source chunks are
independent, so source kernels traverse only prepared subchunks overlapping the
active receiver interval.  Graph growth therefore changes chunk counts and
launches without introducing edge-scaled floating workspace.

\section{Native Channelwise Throughput}
\label{app:uvu_throughput}
\vspace{0.5\baselineskip}
\begin{table}[!htbp]
  \centering
  \caption{Native channelwise \texttt{uvu} capacity-matched results on the fixed 128-node, degree-32 graph. Each pass uses OpenEquivariance's largest graph count $B_{\mathrm{oeq}}$. Paired Peak GiB and M edges/s cells use S $|$ oeq order. Peak GiB is peak allocated CUDA memory. Profiles list multiplicities; $N_{\mathrm w}$ is the radial-weight count.}
  \label{tab:uvu_capacity_matched}
  \scriptsize
  \setlength{\tabcolsep}{2.7pt}
  \renewcommand{\arraystretch}{1.08}
  \resizebox{\linewidth}{!}{%
  \begin{tabular}{@{}lrrrrrrrrrrrrr@{}}
    \toprule
    Profile & $N_{\mathrm w}$ & \multicolumn{4}{c}{\texttt{forward()}} & \multicolumn{4}{c}{\texttt{backward()}} & \multicolumn{4}{c}{\texttt{double backward()}} \\
    \cmidrule(lr){3-6}\cmidrule(lr){7-10}\cmidrule(lr){11-14}
    & & Graphs & Peak GiB & M edges/s & Ratio & Graphs & Peak GiB & M edges/s & Ratio & Graphs & Peak GiB & M edges/s & Ratio \\
    & & $B_{\mathrm{oeq}}$ & S $|$ oeq & S $|$ oeq & S/oeq & $B_{\mathrm{oeq}}$ & S $|$ oeq & S $|$ oeq & S/oeq & $B_{\mathrm{oeq}}$ & S $|$ oeq & S $|$ oeq & S/oeq \\
    \midrule
    $64/64/64/64$ & 1,472 & 1279 & 3.4 | 30.8 & 108.1 | 90.7 & \cellcolor{sobekblue!8}\textbf{1.19$\times$} & 641 & 2.7 | 30.6 & 28.3 | 9.7 & \cellcolor{sobekblue!14}\textbf{2.91$\times$} & 323 & 2.1 | 30.5 & 8.6 | 5.4 & \cellcolor{sobekblue!8}\textbf{1.58$\times$} \\
    \specialrule{0.25pt}{2pt}{2pt}
    $128/128/128$ & 1,408 & 1332 & 3.7 | 30.8 & 390.7 | 86.7 & \cellcolor{sobekblue!22}\textbf{4.51$\times$} & 669 & 3.0 | 30.7 & 41.0 | 9.9 & \cellcolor{sobekblue!22}\textbf{4.15$\times$} & 337 & 2.3 | 30.6 & 9.9 | 5.7 & \cellcolor{sobekblue!8}\textbf{1.74$\times$} \\
    $128/128/128/128$ & 2,944 & 647 & 3.0 | 30.8 & 57.5 | 39.5 & \cellcolor{sobekblue!8}\textbf{1.46$\times$} & 323 & 2.5 | 30.6 & 14.7 | 5.1 & \cellcolor{sobekblue!14}\textbf{2.87$\times$} & 163 & 1.9 | 30.6 & 4.3 | 2.5 & \cellcolor{sobekblue!8}\textbf{1.73$\times$} \\
    \specialrule{0.25pt}{2pt}{2pt}
    $256/256$ & 1,024 & 1807 & 4.5 | 30.8 & 513.8 | 129.1 & \cellcolor{sobekblue!14}\textbf{3.98$\times$} & 909 & 3.6 | 30.6 & 52.5 | 12.5 & \cellcolor{sobekblue!22}\textbf{4.20$\times$} & 460 & 2.8 | 30.6 & 14.9 | 8.0 & \cellcolor{sobekblue!8}\textbf{1.86$\times$} \\
    $256/256/256$ & 2,816 & 672 & 3.4 | 30.7 & 173.4 | 37.0 & \cellcolor{sobekblue!22}\textbf{4.69$\times$} & 336 & 2.8 | 30.6 & 20.2 | 5.3 & \cellcolor{sobekblue!14}\textbf{3.78$\times$} & 169 & 2.1 | 30.5 & 5.0 | 2.6 & \cellcolor{sobekblue!8}\textbf{1.88$\times$} \\
    $256/256/256/256$ & 5,888 & 325 & 2.8 | 30.7 & 28.9 | 13.9 & \cellcolor{sobekblue!14}\textbf{2.07$\times$} & 162 & 2.4 | 30.6 & 6.9 | 2.8 & \cellcolor{sobekblue!14}\textbf{2.50$\times$} & 81 & 1.7 | 30.3 & 2.0 | 1.2 & \cellcolor{sobekblue!8}\textbf{1.74$\times$} \\
    \specialrule{0.25pt}{2pt}{2pt}
    $512/512$ & 2,048 & 914 & 4.1 | 30.8 & 262.2 | 52.7 & \cellcolor{sobekblue!22}\textbf{4.98$\times$} & 458 & 3.4 | 30.7 & 27.4 | 8.6 & \cellcolor{sobekblue!14}\textbf{3.17$\times$} & 231 & 2.5 | 30.5 & 7.8 | 3.9 & \cellcolor{sobekblue!14}\textbf{2.00$\times$} \\
    $512/512/512$ & 5,632 & 338 & 3.2 | 30.8 & 90.5 | 15.4 & \cellcolor{sobekblue!22}\textbf{5.88$\times$} & 168 & 2.7 | 30.5 & 10.6 | 3.0 & \cellcolor{sobekblue!14}\textbf{3.51$\times$} & 85 & 2.0 | 30.5 & 2.6 | 1.3 & \cellcolor{sobekblue!8}\textbf{1.96$\times$} \\
    \specialrule{0.25pt}{2pt}{2pt}
    $1024/1024$ & 4,096 & 459 & 3.8 | 30.7 & 141.1 | 25.8 & \cellcolor{sobekblue!22}\textbf{5.46$\times$} & 229 & 3.3 | 30.6 & 15.0 | 4.1 & \cellcolor{sobekblue!14}\textbf{3.66$\times$} & 116 & 2.4 | 30.6 & 4.2 | 2.0 & \cellcolor{sobekblue!14}\textbf{2.10$\times$} \\
    \bottomrule
  \end{tabular}
  }
\end{table}

\FloatBarrier

\bibliographystyle{unsrt}
\bibliography{references}

\end{document}